\documentclass[lettersize,journal]{IEEEtran}
\usepackage{amsmath,amsfonts}
\usepackage{algorithmic}
\usepackage{array}
\usepackage[caption=false,font=normalsize,labelfont=sf,textfont=sf]{subfig}
\usepackage{textcomp}
\usepackage{stfloats}
\usepackage{url}
\usepackage{verbatim}
\usepackage{graphicx}
\usepackage[colorlinks=true, linkcolor=blue, citecolor=blue, urlcolor=blue]{hyperref}
\usepackage{booktabs}
\def\BibTeX{{\rm B\kern-.05em{\sc i\kern-.025em b}\kern-.08em
    T\kern-.1667em\lower.7ex\hbox{E}\kern-.125emX}}
\usepackage{balance}
\begin{document}
\title{Towards a Reward-Free Reinforcement Learning Framework for Vehicle Control}

\author{Jielong Yang, Daoyuan Huang, Xionghu Zhong, Xin Ge, and Di Wu}


\maketitle
\begin{abstract}
Reinforcement learning plays a crucial role in vehicle control by guiding agents to learn optimal control strategies through designing or learning appropriate reward signals. However, in vehicle control applications, rewards typically need to be manually designed while considering multiple implicit factors, which easily introduces human biases. Although imitation learning methods does not rely on explicit reward signals,  they necessitate high-quality expert actions, which are often challenging to acquire. To address these issues, we propose a reward-free reinforcement learning framework (RFRLF). This framework directly learns the target states to optimize agent behavior through a target state prediction network (TSPN) and a reward-free state-guided policy network (RFSGPN), avoiding the dependence on manually designed reward signals. Specifically, the policy network is learned via minimizing the differences between the predicted state and the expert state, without dependency on expert actions. Experimental results demonstrate the effectiveness of the proposed RFRLF in controlling vehicle driving, showing its advantages in improving learning efficiency and adapting to reward-free environments.
\end{abstract}


\section{Introduction}\label{1}
\IEEEPARstart{I}{n} the field of vehicle control, although traditional control methods such as Model Predictive Control perform well in certain scenarios, they often struggle to achieve ideal control effects when facing dynamically changing environments \cite{10225497,gheisarnejad2022adaptive,10186728}. In contrast, Reinforcement Learning (RL), as a learning-based decision-making method, can learn the policy through the interaction between the agent and the environment \cite{Mnih2013PlayingAW,10020015,zhang2024reinforcement}. The advantage of RL lies in its ability to adapt to the uncertainty and dynamic changes of the environment, learning effective behavioral policies in specific environments through a trial-and-error process \cite{mnih2015human}. Moreover, RL algorithms can automatically adjust and optimize control parameters through the interaction between the agent and the environment \cite{li2024autopilot}. Therefore, researchers are actively exploring various reinforcement learning algorithms and their applications in vehicle control \cite{10547481,shi2023optimal}.

In recent years, reinforcement learning has achieved many remarkable accomplishments in the field of vehicle driving control \cite{chen2023milestones,ye2021survey,grigorescu2020survey}. However, traditional reinforcement learning algorithms, such as the MAD-ARL method \cite{sharif2022adversarial} and the PETS-MPPI method \cite{frauenknecht2023data}, typically require explicit reward signals to guide the vehicle in learning driving policies. Yet, the design of driving rewards for vehicles must consider many implicit factors, making it difficult to define an appropriate reward function for RL \cite{wang2024deep,abouelazm2024review}. Furthermore, in some real-world scenarios, reward signals are very sparse or inaccessible \cite{yang2020adaptive,booher2024cimrl}.

Multiple approaches have been proposed to deal with sparse reward signals. Kamienny et al. \cite{kamienny2020learning} propose the Import algorithm which utilizes informative policies to regularize recurrent neural network training, reducing dependency on explicit rewards, but still requires manually selected task descriptions to learn reward functions. Zintgraf et al. \cite{zintgraf2019varibad} develop the Varibad algorithm, a Bayesian adaptive approach based on meta-learning, which employs a Variational Auto-Encoder (VAE) to learn low-dimensional latent representations of tasks, yet proves ineffective in environments with completely absent rewards. Hejna et al. \cite{hejna2024inverse} introduce the Inverse Preference Learning (IPL) algorithm that enhances stability by simultaneously learning implicit rewards and optimal policies, though it still necessitates manually designed implicit rewards, limiting its application flexibility. Liu et al. \cite{liu2021decoupling} presente the Dream algorithm that identifies task-relevant information by learning exploitation objectives, demonstrating excellent performance in sparse reward scenarios, but requiring additional strategies to ensure effective exploration in completely reward-free environments.

Additionally, when rewards are completely unknown and reward functions are challenging to design, imitation learning emerges as a viable alternative. Imitation learning observes and learns from the behavior of experts, directly extracting behavioral patterns from demonstration data, thus bypassing the reliance on explicit reward signals. Xiao et al. \cite{xiao2023imitation} propose an imitation learning method that enables agents to acquire high-quality behavioral policies by imitating expert behaviors and learning from experience distributions rather than reward functions. Generative Adversarial Imitation Learning (GAIL) \cite{ho2016generative} draws inspiration from Generative Adversarial Network (GAN) \cite{goodfellow2014generative}, training a discriminator to distinguish between expert trajectories and policy-generated trajectories while optimizing the policy to deceive the discriminator. However, imitation learning typically relies on high-quality expert demonstration data, which includes not only state transitions but also the corresponding action information. In some practical applications, obtaining such action information can be challenging because it may involve privacy protection issues or confidentiality of expertise. Generative Adversarial Imitation from Observation (GAIfO) \cite{torabi2018generative}, as a representative method of Imitation from Observation (IfO) \cite{torabi2019recent}, builds upon GAIL and requires only expert state trajectories as input, without the need for expert actions. However, its training process requires fine-tuning the balance between the generator and the discriminator, which can easily lead to training instability or even collapse.

To achieve vehicle control without designing reward functions and without expert actions, we propose a Reward-Free Reinforcement Learning Framework (RFRLF), which updates policies using predicted states and expert states while eliminating dependency on manually designed reward functions. The RFRLF comprises two core components: a Target State Prediction Network (TSPN) and a Reward-Free State-Guided Policy Network (RFSPN). The TSPN predicts the state at the next timestep based on the agent's current state and executed action, and the predicted states is then used to optimize policy network (i.e., RFSPN) without relying on manually designed reward signals. Without the TSPN, the discrepancy between the agent states and expert states cannot directly guide policy learning through backpropagation during the training phase of the RFSPN due to the unknown environment model. Furthermore, the TSPN enables gradual policy network optimization and eliminates the need to simultaneously handle adversarial training dynamics, thereby enhancing learning stability. By training the policy network through loss functions based on state representations rather than manually crafted rewards, we mitigate the human biases introduced by reward function design. Additionally, the RFRLF demonstrates particular suitability for scenarios with unknown expert actions, which are often necessary in conventional imitation learning and used as a substitute for rewards as supervision signals. Experimental results demonstrate the effectiveness of the proposed framework in controlling vehicles.

Compared to existing research, the main contributions of this work are summarized as follows: 
\begin{enumerate}{}{}
	\item{We propose a novel Reward-Free Reinforcement Learning Framework (RFRLF). RFRLF learns agent behavior by observing records of expert demonstrations (such as videos).}
	\item{We design target state prediction networks (TSPN) and reward-free state-guided policy networks (RFSGPN) for the proposed framework. Through their tandem update, policies can be efficiently optimized without reward function design, mitigating performance issues caused by improper reward engineering.}
	\item{This method is particularly suitable for situations where expert actions are unknown, allowing agents to learn effective policies through imitation learning even in the absence of expert actions.}
\end{enumerate}

The rest of this paper is organized as follows. Section \ref{2} will review the research progress in related fields. Section \ref{3} provides a detailed introduction to our proposed reward-free reinforcement learning framework RFRLF, including the design and implementation of the TSPN and the RFSGPN. Section \ref{4} will present our experimental setup, comparison results with baseline methods, and sensitivity analysis of key parameters. Section \ref{5} will summarize the main findings of this study and provide perspectives on future research directions.

\section{Related Work}\label{2}
\subsection{Reinforcement and Imitation Learning in Vehicle Control}
In recent years, researchers have enhanced vehicle control performance through diverse approaches and methodologies \cite{kiran2021deep,li2024iss}. This includes mimicking human driver behavior \cite{sun2023benchmark,gao2019comparison} and mastering intricate, precise policies via reinforcement learning \cite{patel2024enhancing,chen2023risk}.

In \cite{gangopadhyay2022safe}, a method called S2RL is proposed. This method uses Control Barrier Functions (CBFs) and Control Lyapunov Functions (CLFs) to guide the training of the vehicle, ensuring that the learned policy maintains optimal behavior while meeting safety and stability requirements. In \cite{manikandan2023ad}, the authors built the DDQN framework by introducing a dual network architecture, proposing the Dueling Double Deep Q-Network (D3DQN). The D3DQN has shown good performance in both simulation and real-world experiments, effectively navigating vehicles around temporary obstacles and pedestrians. However, these methods typically require explicit reward signals to guide the vehicle's learning to optimize its driving performance. In contrast, our proposed RFRLF method can guide the vehicle to effectively learn driving policies without providing manually designed reward signals. In \cite{gong2024beyond}, a Life-long Policy Learning (LLPL) framework is proposed. This framework combines Imitation Learning (IL) with Life-long Learning (LLL), providing vehicles with a path tracking control policy that can continuously learn and adapt to new environments, significantly improving the learning efficiency and adaptability of the policy. In \cite{zhou2021exploring}, a method based on imitation learning is proposed, which effectively improve the planning and control performance of the vehicle by combining data augmentation, task loss, attention mechanisms, and post-processing planners, achieving robust driving by imitating human driver behavior. However, these methods all require action information in expert data to guide the vehicle's learning of driving policies and optimize performance. In contrast, our proposed RFRLF method learn effective driving policies without providing action information of experts.

\subsection{Reinforcement Learning without Rewards}
In the field of reinforcement learning, when the reward signal is completely missing, the learning process of the agent faces great challenges. To address this challenge, the authors of \cite{ma2024explorllm} propose a method called ExploRLLM, which uses large language models (e.g., GPT-4) to generate common sense reasoning and combines it with the empirical learning ability of reinforcement learning. Specifically, the base model is used to obtain basic strategies, efficient representations, and exploration strategies. By guiding the agent's exploration process through natural language instructions, ExploRLLM can reduce the dependence on explicit reward signals to a certain extent, thereby improving learning efficiency and strategy quality. In \cite{pathak2017curiosity}, the authors propose a curiosity-based exploration mechanism that uses self-supervised learning to predict the consequences of the agent's own actions. Specifically, the agent uses a self-supervised inverse dynamics model to learn the visual feature space, and then uses this feature space to predict state changes caused by the current state and executed actions. In a non-reward environment, curiosity can serve as an intrinsic reward signal, motivating the agent to explore the environment and learn skills that may be useful in the future. In \cite{dawood2023handling}, the authors propose a method that combines Model Predictive Control (MPC) with reinforcement learning to handle non-reward issues. In this method, MPC serves as a source of experience, providing demonstrations to the agent during RL training, thus helping the agent learn how to navigate in non-reward environments. This method successfully improve the agent's learning efficiency and success rate in non-reward environments and reduce the number of collisions and timeouts. In summary, the above methods require manually design of rewards or intrinsic rewards. In contrast, our proposed RFRLF method avoids the manual design of rewards and instead cultivates the decision-making ability of the agent through other mechanisms, thus still demonstrating excellent policy when the design of a reward function is challenging.

\subsection{Control method based on supervised learning of neural network}
It is important to note that although the proposed RFRLF framework incorporates state prediction errors as supervisory signals during training, it fundamentally adheres to the reinforcement learning paradigm. Traditional supervised learning methods primarily rely on static input-output mapping relationships, with their optimization objectives limited to minimizing immediate prediction errors. In contrast, the RFRLF framework generates trajectories through dynamic interaction with the environment and optimizes policies based on the long-term reachability of target states. This interaction-based learning mode preserves the advantages of reinforcement learning in exploration and long-term optimization while avoiding dependence on artificially designed reward functions, which fundamentally distinguishes it from traditional supervised learning methods.

At the implementation level, neural network-based supervised control methods \cite{fei2017adaptive,zhang2020near} typically employ a dual neural network structure. For instance, the dual adaptive neural network control method proposed in \cite{fei2017adaptive}, designed for dynamic control of nonholonomic constrained mobile robots, requires the error between the reference trajectory and current state as input for one of its controllers, and the entire closed-loop control process depends on target state information. In comparison, our proposed RFRLF method uses state errors only as a loss function during the training phase, with the policy network taking only the agent's current state information as input, enabling operation without reliance on expert states during practical application. Furthermore, some traditional control methods require explicit target action information to construct loss functions. For example, \cite{zhang2020near} needs to use MPC to obtain action annotations. For the scenario of learning actions from videos that this paper focuses on, action annotations are difficult to obtain. The RFRLF method, however, can learn effective control policies without any action annotation information, significantly expanding its range of applications.

\section{Methodology}\label{3}

\begin{figure*}[!t]
	\centering
	\includegraphics[width=0.9\textwidth]{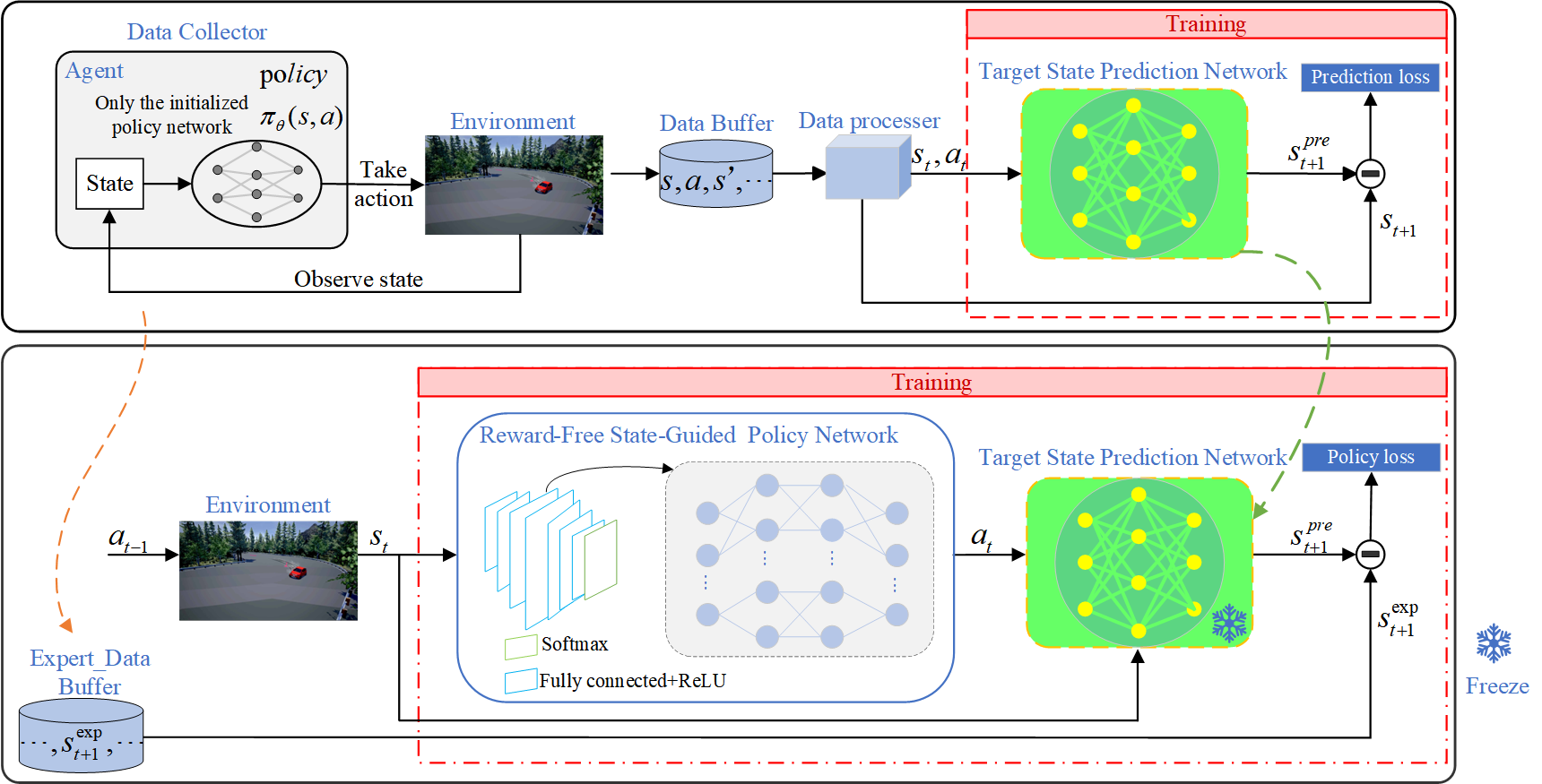}
	\caption{The architecture of our reward-free reinforcement learning framework, which consists of two parts. In the first part, we collect state-action pairs and train a target state prediction network. In the second part, we train the reward-free state-guided policy network using the predicted states provided by the target state prediction network along with the state data provided by the expert, while freezing the target state prediction network.}
	\label{fig:Overall_framework}
\end{figure*}

In practical applications, designing effective reward functions is often challenging and time-consuming, especially in environments with sparse or ambiguous rewards. To address the challenge of reward signal dependency in traditional reinforcement learning, this chapter details our proposed Reward-Free Reinforcement Learning Framework (RFRLF), as illustrated in Fig. \ref{fig:Overall_framework}. We divide the framework into two complementary modules: the Target State Prediction Network (TSPN) and the Reward-Free State-Guided Policy Network (RFSGPN). The TSPN predicts the next state based on the agent's current state and actions, providing a clear target information for the policy network. The RFSGPN utilizes the target state predictions and expert state information to optimize the agent's policy, enabling effective learning in environments without explicit rewards. These two modules work in tandem; the TSPN supplies the necessary target information, and the policy network uses this information to refine actions, thereby reducing reliance on manually designed reward signals and enhancing adaptability and learning efficiency in complex environments. Distinguished from neural network-based supervised control methods \cite{fei2017adaptive,zhang2020near}, the proposed method offers the following advantages: 
\begin{enumerate}{}{}
	\item{It uses expert states to compute state errors as loss functions only during the training phase, requiring only the agent's current state information during practical application, without dependence on expert states.}
	\item{It can learn effective control policies without action annotation information, greatly expanding its application scope, particularly suitable for scenarios such as learning actions from videos.}
\end{enumerate}
In the following sections of this chapter, we will delve into the specific implementations of these two modules and how they collectively address the challenges (i.e.,dependence on manually designed rewards or expert actions) in reinforcement learning.

\subsection{Target State Prediction Network (TSPN)}
The design of the TSPN stems from an in-depth understanding of environmental dynamics, aiming to predict the state at the next time step by modeling the interactions between the agent and the environment, thereby providing predictions of target states for subsequent policy optimization. This approach circumvents the reliance on manually designed reward signals in traditional reinforcement learning, enabling the policy network to directly learn how to achieve target states. This network uses random interaction data $(s_t,a_t)$ from the agent and the environment to predict the target state $s_{t+1}^{pre}$, and the predicted target state is used alongside the next state $s_{t+1}$ from the environmental interaction to optimize the prediction network. The optimized prediction network is then used to train the RFSGPN, which serves as the policy network. This part of the content will be introduced in the following section \hyperref[sec:reward-free state-guided policy network]{III-B}.

To provide the TSPN with continuous and diverse state-action pair observation data, we design a data collector and a data processor. As shown in Fig. \ref{fig:Overall_framework}, the data collector uses an untrained policy network to interact randomly with the environment to obtain observational data. During this process, a series of (state, action, next state) triplet data generated are saved to the data buffer. Subsequently, the data processor processes the data in the data buffer to ensure that the extracted observation sequences are continuous and do not contain end flags. This process ensures the quality of the training data for the TSPN, enabling it to learn on diverse state-action sequences.

\begin{figure*}[!t]
	\centering
	\includegraphics[width=0.9\textwidth]{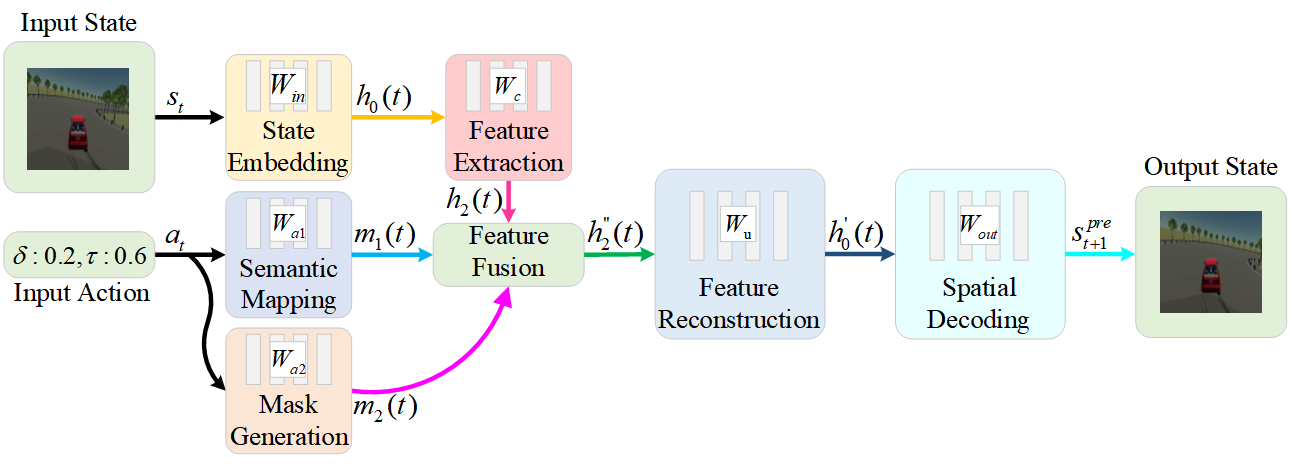}
	\caption{The structure of the target state prediction network (TSPN). In these networks, the input state is first processed by the input layer and then passed through the feature extraction layer to extract key state features. Further, the action information is fused into these state features through the action injection layer. Finally, through the feature reconstruction and the spatial decoding layer, the network generates a prediction of the target state.}
	\label{fig:Target_prediction_network}
\end{figure*}

Based on the high-quality training data, the TSPN takes the current state and action information as inputs and outputs the prediction of the state at the next time step. As shown in Fig. \ref{fig:Target_prediction_network}, the TSPN is composed of multiple modules, including the input embedding layer, feature extraction layer, feature reconstruction layer, action injection layer (including semantic mapping, mask generation and feature fusion), and the spatial decoding layer. These modules not only effectively capture the features of the current state but also consider the agent's decisions and actions through the action injection layer, enabling the network to better capture the correlation between the previous and subsequent states as well as the actions. In the following sections, we will provide a detailed introduction to these modules.

The input embedding layer uses a convolutional neural network to embed the input state frames. We select appropriate kernel sizes and strides to ensure that key information in the state frames can be effectively extracted, thereby improving the accuracy of predictions. We design two feature extraction layers that reduce the dimension of features obtained after embedding through multiple convolutions and instance normalization operations, reducing computational complexity and memory consumption and allowing the network to consider contextual information over a larger range. Specifically, the input state is processed as follows:
\begin{equation}
	\begin{gathered}
		h_0(t)=\text{ReLU}(W_{in}s_t+b_m),\\
		h_1(t)=\text{InstanceNorm}(\text{ReLU}(W_{c1}h_0(t)+b_{c1})),\\
		h_2(t)=\text{InstanceNorm}(\text{ReLU}(W_{c2}h_1(t)+b_{c2})),
	\end{gathered}
\end{equation}
where $W_{in}$, $b_{in}$, $W_{c1}$, $b_{c1}$, $W_{c2}$ and $b_{c2}$ are the weight matrices and bias vectors of the network, respectively. 

On this basis, the action injection layer injects action information into the feature representation, enabling the network to make state predictions based on action information. In reinforcement learning tasks, the next state depends not only on the current state but also on the actions taken. Therefore, injecting action information during the feature extraction and feature reconstruction of the network can help the network better understand the impact of actions on state changes, thereby improving the accuracy of predictions. The action injection layer employs dual-branch fully connected layers to generate action masks: the Mask Generation branch maps actions to a channel attention mask $m_1(t)$ , whose elements are either 0 or 1, and the Semantic Mapping branch linearly transforms actions into a compensation mask $m_2(t)$. During feature fusion, the layer first performs element-wise multiplication between the state feature $h_2(t)$ and $m_1(t)$, then applies residual compensation by adding $m_2(t)$, ultimately producing enhanced features with integrated action information. This process is formulated as follows:
\begin{equation}
	\begin{aligned}
		m_1(t) &= \sigma(W_{a1}a_t),\\
		h_2^{'}(t) &= h_2(t) \odot m_1(t),\\
		m_2(t) &= W_{a2}a_t,\\
		h_2^{''}(t) &= h_2^{'}(t) + m_2(t),
	\end{aligned}
\end{equation}
where $a_t$ is the current action, $W_{a1}$, $W_{a2}$ are the weight matrices of the action injection layer, $\sigma$ is the \text{sigmoid} function, and $\odot$ represents element-wise multiplication. Subsequently, the feature reconstruction layer reconstructs the feature representation through multiple deconvolution and instance normalization operations. We designe two feature reconstruction layers to map the features after action injection layer, $h_2^{''}(t)$, back to a higher-dimensional representation $h_1^{'}(t)$ and the final embedded representation $h_0^{'}(t)$ in sequence:
\begin{equation}
	\begin{gathered}
		h_1^{'}(t)=\text{InstanceNorm}(\text{ReLU}(W_{u1}h_2^{''}(t)+b_{u1})),\\
		h_0^{'}(t)=\text{InstanceNorm}(\text{ReLU}(W_{u2}h_1^{'}(t)+b_{u2})).
	\end{gathered}
\end{equation}
Finally, the spatial decoding layer uses a convolutional neural network to spatially decode the reconstructed features and obtain the prediction of the next state. This design allows the network to directly output pixel-wise prediction results, which is formulated by 
\begin{equation}
	s_{t+1}^{pre}=W_{out}h_0^{'}(t)+b_{out}.
\end{equation}

During the training process, our goal is to make the network's prediction of the next state $s_{t+1}^{pre}$ as close as possible to the actual next state $s_{t+1}$. We choose to use the mean squared loss function to train the network. Specifically, the definition of the loss function is as follows:
\begin{equation}
	\mathcal{L} =\frac{1}{N}\sum_{i=1}^{N}(s_{t+1}^{pre}-s_{t+1})^2,
\end{equation}
where $N$ is the batch size.

It is worth noting that when the input to the TSPN consists of numerical data, convolutional layers are not required for prediction. Therefore, in this case, we replace the original convolutional and deconvolutional layers with fully-connected layers. Additionally, the instance normalization function is replaced by a layer normalization function, which is more appropriate for fully-connecteds layers.

In summary, the TSPN processes the interaction data between the agent and the environment to generate predictions of future states. These predictions serve not only to refine the network itself but also provide crucial target state guidance for the policy network. In the following section \hyperref[sec:reward-free state-guided policy network]{III-B}, we will explore how to use these predicted states in combination with expert state data to learn the RFSGPN.

\subsection{Reward-Free State-Guided Policy Network (RFSGPN)}\label{sec:reward-free state-guided policy network}
In practice, explicit reward signals are often difficult to obtain. Even when reward signals can be acquired, their effectiveness may be compromised by noise, delay, or design complexity, which in turn affects the learning of the policy \cite{yang2020adaptive}, \cite{chen2021self}. To address this issue, we design a RFSGPN. The motivation of this network is to use the predicted state as a substitute supervision signal, directly providing the agent with learning objectives. By leveraging predicted target states and state information provided by experts, the network continuously guide the learning process without the need for manually designed rewards and in the absence of immediate reward feedback, achieving effective policy optimization.

Next, we introduce the design and implementation of the RFSGPN. The RFSGPN extracts state features through a network $Q(\cdot)$ composed of convolutional layers, batch normalization, and fully connected layers, which is formulated by 
\begin{equation}
	f_t=Q(s_t),
	\label{eq:Q}
\end{equation}
where $s_t$ is the current input state, initialized as the provided initial state, and subsequent states are obtained as the next state after interaction with the environment.\cite{zhang2022traffic}. When the input of RFSGPN is numerical data, it is not suitable to use convolutional layers to extract features. we change the convolution layers in \eqref{eq:Q} to fully-connected layers. Subsequently, a fully connected layer with the Gumbel-Softmax function generates the action probability distribution. For most reinforcement learning algorithms, since the environment dynamics are unknown, the discrepancy between the environment-generated states and target states cannot directly guide policy learning through backpropagation. However, by leveraging our pre-trained prediction network, states can propagate gradients through this network to optimize the policy. We therefore introduce a novel mechanism: directly feeding the policy distribution $\pi(a_t\mid s_t)$ into the parameter-frozen Target State Prediction Network (TSPN) to predict the next state $s_{t+1}^{pre}$, then using the discrepancy between predicted and target states for policy optimization via backpropagation. This approach eliminates both the need for explicit action sampling and the requirement for value networks, fundamentally circumventing dependency on reward engineering. The process is formally expressed as:
\begin{equation}
	s_{t+1}^{pre}=\text{TargetNet}_{\text{frozen}} (s_t,\pi(a_t \mid s_t)),
\end{equation}
where $\text{TargetNet}_{\text{frozen}}$ represents the input embedding and feature extraction layers of the TSPN with frozen parameters, and $s_t$ represents the environment-generated state at each timestep. Here we use actual environment states $s_t$ rather than predicted states, which helps correct errors in the prediction model during policy learning. To obtain $s_t$ at each step, actions $a_t$ are sampled from $\pi(a_t\mid s_t)$ and this process does not participate in backpropagation. For discrete action spaces, direct sampling is employed, whereas continuous action spaces require discretization prior to sampling. These implementation specifics are omitted here as they are non-essential to the core contributions of this work. After obtaining $s_{t+1}^{pre}$, we compute the following loss function to train the policy network: 
\begin{equation}
	\mathcal{L} =\frac{1}{N}\sum_{i=1}^{N}(s_{t+1}^{pre}-s_{t+1}^{exp})^2,
\end{equation}
where $N$ is the batch size, $s_{t+1}^{exp}$ is the next state of the expert data. In this way, our network training process leverage the advantage of TSPN in state prediction and advantage of RFSGPN in policy learning, achieving effective learning without manually designing reward signals. Hence, our method is applicable to environments where reward signals are difficult to design or missing.

\section{Experiments and Discussion}\label{4}
This section elaborates on the experimental setup of our method, covering key steps such as data collection, data processing, and network training. Our method is benchmarked against two existing reward-free and two reward-based reinforcement learning methods in the Carla and Autocar environments. Additionally, a sensitivity analysis of key parameters further validates the robustness and generalization capability of our approach. The experiments are conducted to answer the following questions:
\begin{itemize}{}{}
	\item{Can RFRLF effectively optimize the agent’s behavior without manually designing reward signals?}
	\item{When using only expert state information, can the joint TSPN and the RFSGPN enable the agent to learn effective policies?}
	\item{How do different parameter settings affect the performance of the algorithm, and can it maintain its effectiveness under different training conditions?}
\end{itemize}

\subsection{Experimental Setup}
In this part, we introduce our methods for data collection and processing. Subsequently, we describe the training methodology for the network. 

\subsubsection{Data Collection}
The data collection is executed by a data collector equipped with an initialized policy network. It engages in random interactions with the environment, collecting observational data and predicting actions through an untrained, initialized policy agent. The purpose of this process is to gather a series of (state, action, next state) triplets, which will be used for the training of TSPN.

During the data collection process, we use a temperature parameter to adjust the randomness of the actions selected by the agent. We set the temperature parameter to 1.0 to maintain a balance between exploration and exploitation. The setting of this parameter is an important experimental factor, which we study in detail in the subsequent section \hyperref[sec:Parameter sensitivity analysis]{IV-F}.

\subsubsection{Data processing}
Data processing is carried out after the data collection is completed. We designe a data processor that extracts data of the set batch size from the collected data buffer. During the data processing, the data processor ensures that the extracted observational sequences do not contain end flags. The design of this mechanism is to ensure the quality of the training data, enabling the network to learn on meaningful state sequences. 

\subsubsection{Network Training Configuration}
In our framework, training is divided into two phases: independent training of the TSPN followed by the RFSGPN training, with the latter conducted while keeping the parameters of the former fixed. This section details the training parameter settings and specific implementations for both phases.

For the TSPN, we employ the Adafactor optimization algorithm with an initial learning rate of 0.001 using a cosine annealing strategy, a batch size of 64, and train for 100 epochs, each containing 1000 iterations. We utilize mean squared error as the loss function and implement Scheduled Sampling with an initial $\epsilon$ value of 0.9, which linearly decays to 0.5 during the first two training cycles and remains constant thereafter. This configuration enables the network to primarily rely on actual state sequences during the initial training phase, gradually transitioning to utilizing its own prediction results, effectively enhancing model stability.

For the RFSGPN training, we also utilize the Adafactor optimization algorithm but with a fixed learning rate of 0.0005, a batch size of 32, and train for 50 epochs with an early stopping strategy based on validation set performance. During the training process, we fix the parameters of the pre-trained TSPN and only update the policy network parameters, performing gradient backpropagation by computing the mean squared error between predicted states and expert states. The entire training process is implemented using the PyTorch 1.10.0 framework on an NVIDIA RTX 3090 GPU, with gradient clipping (maximum gradient norm of 1.0) to ensure training stability.

\subsection{Baselines}
In our experiments, we compare the proposed RFRLF method with four different baseline methods: Import \cite{kamienny2020learning}, Varibad \cite{zintgraf2019varibad}, IPL \cite{hejna2024inverse}, and Dream \cite{liu2021decoupling}.

Import: An algorithm proposed by Kamienny et al. \cite{kamienny2020learning}. While not completely reward-free, Import reduces reliance on explicit rewards by using informed policies to regularize the training of recurrent neural network policies.

Varibad: A Bayesian adaptive deep reinforcement learning method based on meta-learning proposed by Zintgraf et al. \cite{zintgraf2019varibad} uses a Variational Auto-Encoder (VAE) to learn a low-dimensional stochastic latent representation of tasks. Through meta-learning and modeling of task uncertainty, the agent can effectively explore and exploit in an environment with unknown rewards.

IPL: A reward-free algorithm proposed by Hejna et al. \cite{hejna2024inverse}, designed to learn policies through preference-based reinforcement learning without the need to explicitly learn a reward function. IPL utilizes offline preference data and optimizes a parameter-efficient algorithm to directly learn policies from human feedback.

Dream: A meta reinforcement learning framework proposed by Liu et al. \cite{liu2021decoupling}. This framework avoids the problem of local optimum by optimizing exploration and exploitation separately, allowing the agent to learn quickly on new tasks. Notably, it does not depend on explicit reward signals, thereby selected as a baseline of our method.

\subsection{Simulation Environmental Experiment Analysis}
\subsubsection{Carla Environment Experiment Analysis}
The environment is created by Cai et al. \cite{cai2020high} in the high-fidelity driving simulator Carla \cite{dosovitskiy2017carla}, which provides a highly realistic virtual environment for training and evaluating autonomous driving policies. The map is generated by RoadRunner \cite{vectorzero2019roadrunner}, a road and environment creation software used for vehicle simulation. Our goal is to ensure that the vehicle travels forward without collision along a small number of reference trajectories within the map's lanes. The entire training and testing process is conducted in Carla.

\textbf{States and Actions:} The state space of the environment is defined as follows:
\begin{equation}
	S=\{\delta,\tau,e_y,\dot{e}_y,e_{\varphi},\dot{e}_{\varphi},e_{\beta},\dot{e}_{\beta},e_{vx},\dot{e}_{vx},e_{vy},\dot{e}_{vy},T\},
\end{equation}
where $\delta$ is the steering angle of the vehicle; $\tau$ is the throttle size; $e_y$ is the cross track error, defined as the perpendicular distance of the vehicle from the reference track; $e_{\varphi}$ is the heading angle error, which is the difference between heading angle of the vehicle and the desired heading angle; $e_{\beta}$ is the slip angle difference between the vehicle and the reference trajector; $e_{vx}$ and $e_{vy}$ are the longitudinal and lateral velocity errors, respectively; $\dot{e}_y, \dot{e}_{\varphi}, \dot{e}_{\beta}, \dot{e}_{vx}$ and $\dot{e}_{vy}$ are the time derivatives of $e_y, e_{\varphi}, e_{\beta}, e_{vx}$ and $e_{vy}$, respectively; $T$ contains ten $(x,y)$ positions and slip angle in the reference trajectories ahead. Thus, the dimension of $S$ is 42. The action space is defined as follows:
\begin{equation}
	A=\{\delta,\tau\}.
\end{equation}
Here we limit the vehicle's steering angle $\delta$ and throttle $\tau$ to the range of [-0.8, 0.8] and [0.6, 1], respectively.

\textbf{Rewards:} We define the reward as \eqref{eq:r}, which is the product of the vehicle speed and the weighted sum of the partial rewards:
\begin{equation}
	r=v(40r_{e_y}+40r_{e_{\varphi}}+20r_{e_{\beta}}),
	\label{eq:r}
\end{equation}
where $r_{e_y}$, $r_{e_{\varphi}}$, $r_{e_{\beta}}$ are defined as follows:
\begin{equation}
	\begin{gathered}
		r_{e_y}=e^{-k_1e_y},\\
		r_{e_{\varphi}}=
		\begin{cases}
			e^{-k_1|e_{\varphi}|}& |e_{\varphi}|<90^{\circ}\\
			-e^{-k_2(180^{\circ}-e_{\varphi})}& e_{\varphi}\geq90^{\circ}\\
			-e^{-k_2(180^{\circ}+e_{\varphi})}& e_{\varphi}\leq90^{\circ}
		\end{cases},\\
		r_{e_{\beta}}=
		\begin{cases}
			e^{-k_1|e_{\beta}|}& |e_{\beta}|<90^{\circ}\\
			-e^{-k_2(180^{\circ}-e_{\beta})}& e_{\beta}\geq90^{\circ}\\
			-e^{-k_2(180^{\circ}+e_{\beta})}& e_{\beta}\leq90^{\circ}
		\end{cases}.
	\end{gathered}
\end{equation}

%

\noindent Among them, $k_1$ and $k_2$ take the values of 0.5 and 0.1, respectively. Although we define rewards, they are only used to verify the effectiveness of our method in the final testing phase and are not utilized during the training process. Additionally, when the vehicle collides with an obstacle, reaches the destination, or is more than 15 meters away from the track, 'done' becomes true, ending the current episode.

\begin{figure}[!t]
	\centerline{\includegraphics[width=3.15in,height=1.98in]{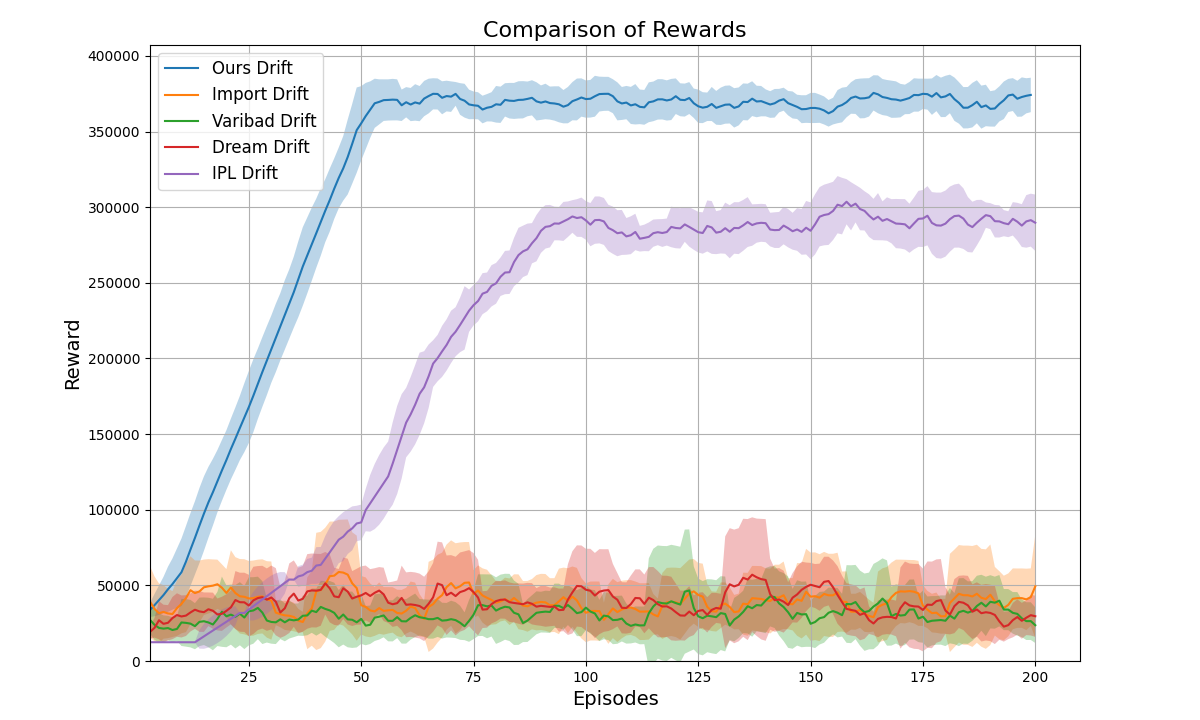}}
	\caption{Test results of different methods on the Carla environment.}
	\label{fig:Carla environmental test results}
\end{figure}

\textbf{Evaluation:} Here, we compare the performance of our method with four baseline methods. We evaluate the performance of the agent under each method for 200 episodes and present the testing performance of different methods on this task in Table \ref{Table1}. Correspondingly, Fig. \ref{fig:Carla environmental test results} shows the test results for each episode during the testing period. In Table \ref{Table1}, in addition to the maximum and average values, we also use the Inter-Quartile Mean (IQM) and the normalized IQM as evaluation metrics. These two indicators can effectively reduce the impact of extreme values on the average, thereby more accurately reflecting the central tendency of the test results. Table \ref{Table1} and Fig. \ref{fig:Carla environmental test results} indicate that, overall, RFRLF (Ours) has surpassed existing baselines in terms of maximum, average, and IQM metrics.

\begin{table}[htbp]
	\centering
	\caption{Evaluation Results of Different Methods in the Carla Environment Experiment, Including Maximum, Average, and IQM Metrics.}
	\label{Table1}
	\begin{tabular}{p{1.5cm}<{\centering}p{1.2cm}<{\centering}p{1.2cm}<{\centering}p{1.2cm}<{\centering}p{1.2cm}<{\centering}}
		\toprule
		\textbf{Method} & \textbf{Max Value} & \textbf{Mean Value} & \textbf{Original IQM} & \textbf{Normal IQM} \\
		\midrule
		Ours & 3.97$\times$10\textsuperscript{5} & 3.29$\times$10\textsuperscript{5} & 3.64$\times$10\textsuperscript{5} & 0.91 \\[0.2cm]
		IPL(2023) & 3.20$\times$10\textsuperscript{5} & 2.16$\times$10\textsuperscript{5} & 2.56$\times$10\textsuperscript{5} & 0.79 \\[0.2cm]
		Dream(2020) & 1.47$\times$10\textsuperscript{5} & 3.86$\times$10\textsuperscript{4} & 3.47$\times$10\textsuperscript{4} & 0.16 \\[0.2cm]
		Import(2020) & 1.32$\times$10\textsuperscript{5} & 3.93$\times$10\textsuperscript{4} & 3.41$\times$10\textsuperscript{4} & 0.18 \\[0.2cm]
		Varibad(2019) & 1.33$\times$10\textsuperscript{5} & 3.07$\times$10\textsuperscript{4} & 2.62$\times$10\textsuperscript{4} & 0.15 \\
		\bottomrule
	\end{tabular}
\end{table}

From the results, it can be seen that the Import, Varibad and Dream algorithms do not achieve ideal performance. Although the Import algorithm reduces the dependence on the reward signal to a certain extent by utilizing the information of the task description, it is difficult to find an effective strategy based on this alone when the reward signal is completely missing. The Varibad algorithm involves a complex VAE model and meta-learning process, which may not be able to adapt to the reward-free environment and the autonomous driving scenario that requires rapid response, resulting in decision delays and failure to learn effective strategies. In the absence of such a reward signal, the Dream algorithm may require additional information or policies for effective exploration and learning. In addition, although IPL can simultaneously learn implicit rewards and optimal policies that meet the expert's preferences, and can improve the learning process by collecting feedback data, it has a relatively good effect in this task, but expert preference data still introduce human errors. Compared with these methods, the RFRLF framework we proposed shows significant advantages. RFRLF optimizes the behavior of the agent and guides it to learn an effective policy without manually designing reward signals by combining the TSPN and the RFSGPN.

\subsubsection{Autocar Environment Experiment Analysis}
\begin{figure*}[!t]
	\centering
	\includegraphics[width=0.8\textwidth]{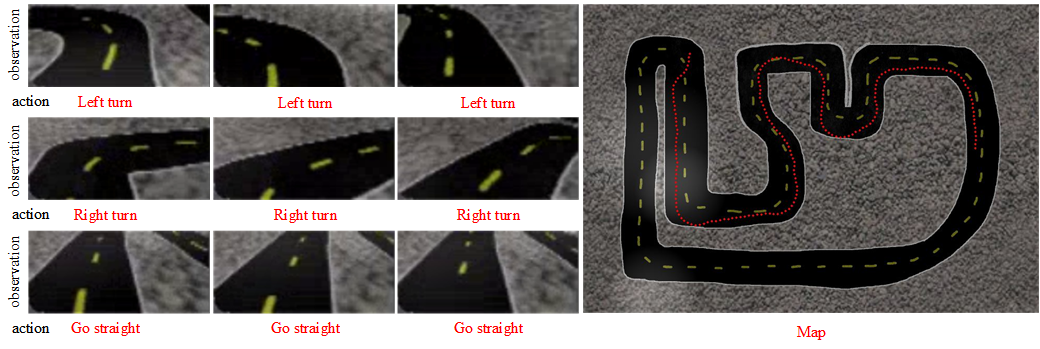}
	\caption{A sequence diagram of state-action pairs (time from left to right) for some vehicles when turns left, turns right, and goes straight  and a map of the environment. On the left side of the figure, the upper row shows the first-person perspective (i.e., states) of the vehicle during driving, and the lower row shows the actions taken by the vehicle in these states. On the right side of the figure, a complete map of the driving environment is shown, showing the approximate position and path of the vehicle in the simulated environment.}
	\label{fig:Autocar environment picture}
\end{figure*}

This environment is created by Maxime Ellerbach \cite{ellerbach2019rl} and is designed to simulate real-world driving scenarios. As shown on the left side of Fig. \ref{fig:Autocar environment picture}, in this environment, the agent drives from a first-person perspective with a limited field of view and is unable to obtain global information. The agent can only observe a small area in front and around it, which makes the decision-making process more complicated. As shown on the right side of Fig. \ref{fig:Autocar environment picture}, the main goal of the agent (indicated by the red dot) is to avoid collisions with road boundaries or other obstacles while maintaining safety, while driving as far forward as possible to increase the driving distance.

\textbf{States and Actions:} The state consists of three-dimensional image data, which reflects the observation of the vehicle in its limited field of view. The agent can perform three basic actions: turn left, go straight, and turn right. The left side of Fig. \ref{fig:Autocar environment picture} shows the timing diagram of several groups of vehicles taking corresponding actions in different states. Through these images, we can observe how the agent chooses appropriate actions to avoid collisions and optimize the driving path when facing different environmental conditions. For example, in a state where a turn is required, the agent may choose to turn left or right; while on a wide straight road, it may choose to go straight.

\textbf{Rewards:} If the vehicle is moving forward, the reward is the current speed; if the vehicle is turning, the reward is half of the current speed (to prevent excessive turning); if the vehicle is making a half-circle turn, the reward is -45; if the vehicle goes off the track or has a collision, the reward is -150, and the episode ends. Like the previous experiment, although we define rewards, they are only used to verify the effectiveness of our method in the final testing phase and are not used in the training process.

\textbf{Evaluation:} Here, we assess and compare the performance of our proposed RFRLF method against four baseline methods. In our experiments, we evaluate the performance of  the agent under each method in 200 episodes. The test results show that the RFRLF method achieves the same performance as the IPL algorithm, Varibad algorithm, and Import algorithm among the four baseline algorithms, all achieving a test reward of 414. This score is significantly higher than the rewards obtained by the Dream algorithm.

The results are due to two reasons. First, the Dream algorithm has limited feature extraction capabilities when processing image data, making it difficult to extract sufficiently accurate features from it and unable to adapt well to the nuances of the environment. In contrast, the TSPN in our proposed RFRLF framework is able to extract state features from image data and combine the corresponding action features with state features to predict the next state. In addition, RFSGPN has a specialized feature extraction network that can effectively process environmental state features, thereby better understanding the environment and adjusting policies. Second, although the Dream algorithm performs well in processing sparse rewards, it has difficulty effectively balancing exploration and exploitation in environments where rewards are completely lacking, probably because it requires sparse rewards to evaluate the utility of its exploration. In contrast, the policy optimization process of RFRLF does not dependent on manually designed reward feedback, but is instead guided by the loss function of the state representation, providing a more stable supervision signal for the learning process in a reward-free environment.

\subsection{Real-world Scenario Experimental Validation}
To verify the effectiveness of the proposed RFRLF framework in real-world environments, this section constructs an actual experimental platform based on the TurboPi intelligent vision vehicle and conducts tests on an autonomous driving track. The experimental objective is to validate whether RFRLF can achieve stable control solely through expert state information without reward signals and expert action guidance.

\textbf{Hardware Platform:} The experiment employs the TurboPi intelligent vision vehicle, with core components including a Raspberry Pi 4B control board, an RGB camera (resolution 640×480), a 4-channel line-tracking sensor, an LFD-01 PMM servo, and a Mecanum wheel drive system. The trained policy network is deployed on the Raspberry Pi 4B, and control commands are inferred in real-time using Python.

\textbf{Expert Data:} The vehicle is guided to autonomously drive along the track centerline 10 times using the 4-channel line-tracking sensor, collecting continuous state trajectories (including camera images) as expert state inputs.

\begin{figure*}[!t]
	\centering
	\includegraphics[width=0.8\textwidth]{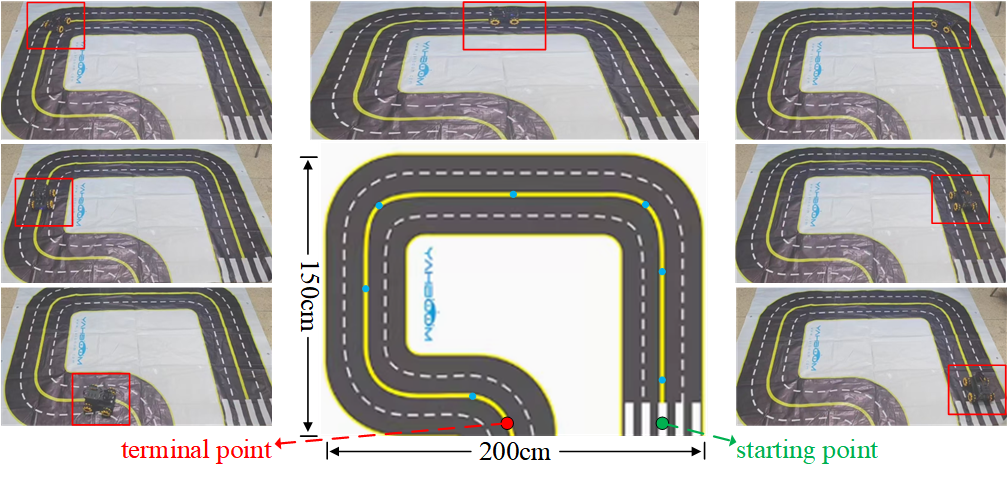}
	\caption{The central area is a 200×150 cm autonomous driving track, with the starting point and endpoint marked. Surrounding the track are top-down views of the vehicle driving at different positions, corresponding to the blue marking points on the track.}
	\label{fig:Actual test track display diagram}
\end{figure*}

\textbf{Experimental Environment:} As shown in Fig. \ref{fig:Actual test track display diagram}, the track is 200cm $\times$ 150 cm, including straight and curved sections.

\textbf{Experimental Results:} Fig. \ref{fig:Actual test track display diagram} illustrates the actual experimental scenario layout and the vehicle's driving states at multiple positions under RFRLF control. The experimental results demonstrate that, in the absence of reward signals and expert actions, RFRLF can optimize the policy by predicting the error between the state and the expert state, enabling the vehicle to drive stably along the track centerline with an average lateral error of 3.2 cm (standard deviation ±1.5 cm).

\subsection{Performance Evaluation of Target State Prediction Network}
The TSPN is a key component of the RFRLF framework, which aims to predict the next state of the environment. This TSPN provides prediction values for the RFSGPN, which is compared with the state information of the expert to help the agent learn effective policies in the reward-free environment. This section takes the Autocar environment task as an example to evaluate the performance of the TSPN and show the accuracy of its prediction.

\begin{figure}[!t]
	\centering
	\begin{minipage}{0.45\textwidth}
		\centering
		\includegraphics[width=\linewidth]{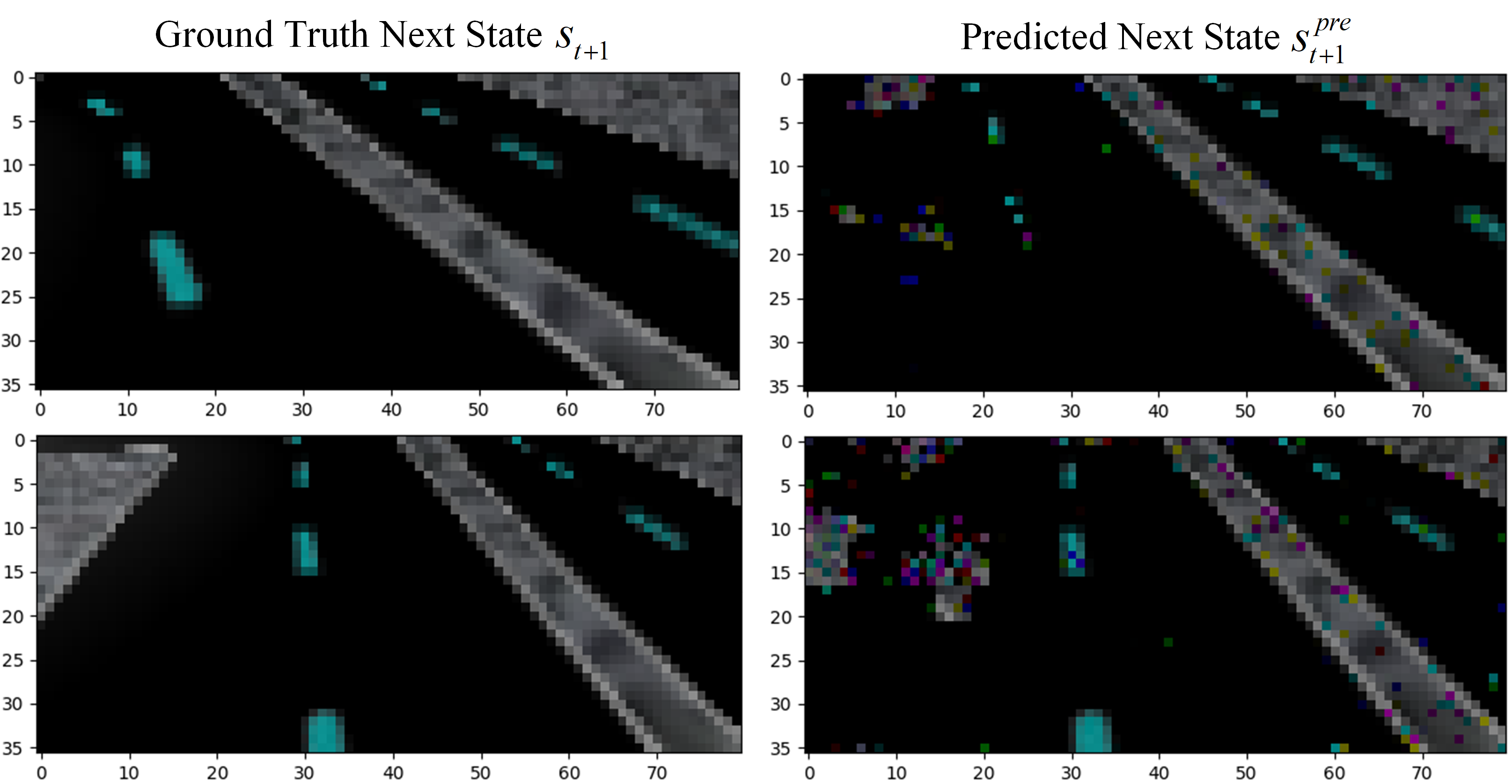}
		{\footnotesize (a)}
	\end{minipage}
	\hfill
	\begin{minipage}{0.45\textwidth}
		\centering
		\includegraphics[width=\linewidth]{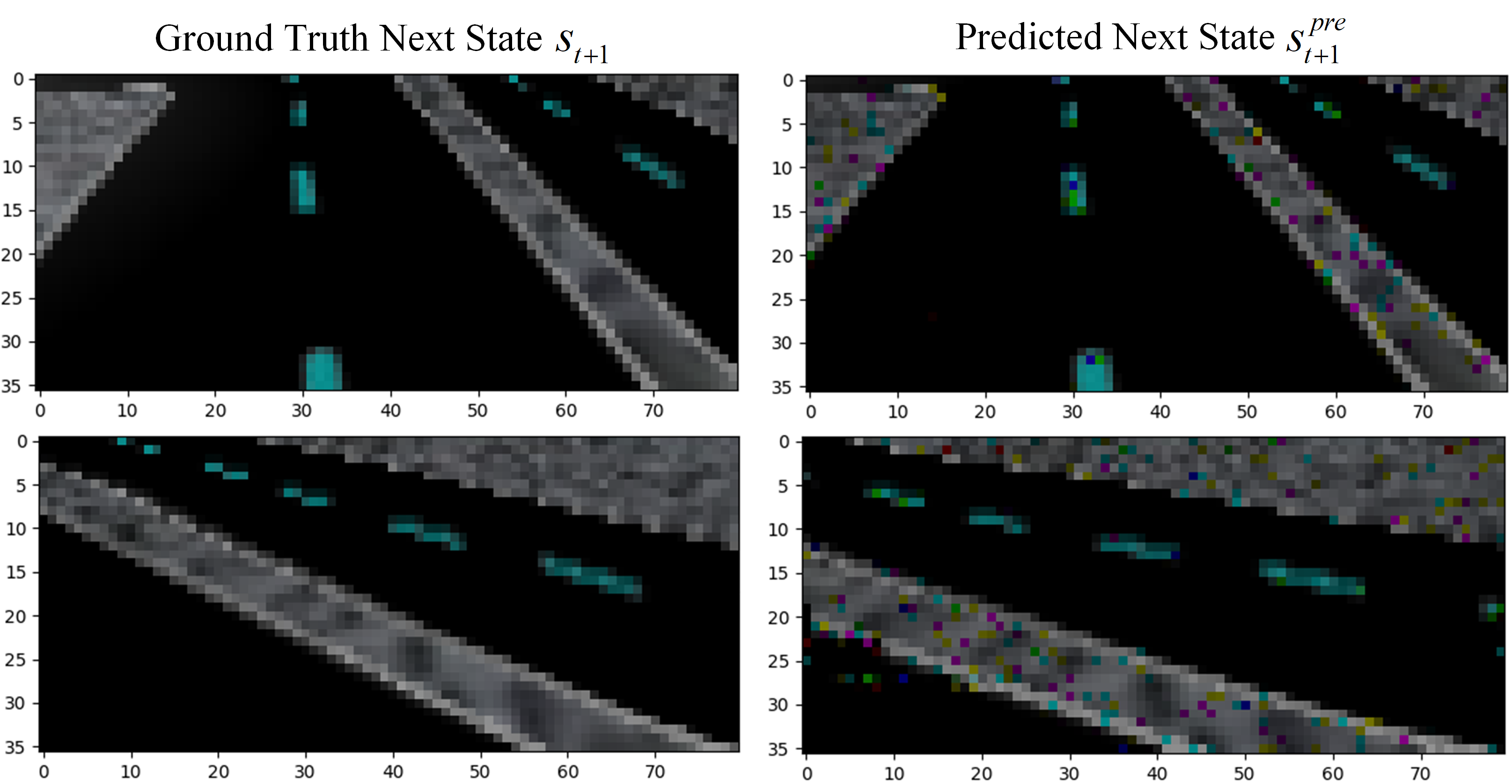}
		{\footnotesize (b)}
	\end{minipage}
	\caption{Prediction performance of the TSPN after training. (a) Prediction performance of the TSPN after 5 epochs of training. (b) Prediction performance of the TSPN after 10 epochs of training.}
	\label{fig:Comparison of prediction model performance}
\end{figure}

We train the TSPN for 10 epochs. The specific network configuration and hyperparameters are as follows: the size of the state frame is 3×159×255, the batch size is 16, and the learning rate scheduler is set to decay the learning rate after every 5,000 steps, with a decay coefficient of 0.5. In order to verify the performance improvement of the target state prediction network, we re-collect test data after every 5 epochs of training, and use this data to test the TSPN, comparing the next state $s^{pre}_{t+1}$ predicted by the network with the actual next state $s_{t+1}$. Fig. \hyperref[fig:Comparison of prediction model performance]{6} shows the effect of the network after training. It can be clearly seen from the figure that as the training progresses, the prediction accuracy of the network is significantly improved. In particular, in Fig. \hyperref[fig:Comparison of prediction model performance]{6 (b)}, the predicted state generated by the TSPN is very close to the actual state, indicating that the network has been able to effectively learn the relationship between the current state and the action, and predict the dynamic changes of the environment.

The key reason why the TSPN can achieve high-precision predictions lies in the action injection layer in its design. By injecting action information into the state features, the network is able to capture the impact of actions on state changes. In addition, the feature reconstruction and feature extraction modules not only effectively preserve the spatial resolution of the features, but also enable the network to capture richer contextual information, thereby further improving the prediction accuracy.

In general, the TSPN performed well in the experiment and is able to accurately predict the next state of the environment, which is crucial for the RFRLF framework. Accurate state prediction provides a reliable foundation for the training of the RFSGPN, allowing the RFSGPN to optimize the behavior of the agent and learn effective policies even in the absence of manually designed reward signals.

\subsection{Parameter Sensitivity Analysis}\label{sec:Parameter sensitivity analysis}
In the Carla environment, we implement 5 comparison schemes by setting various hyperparameters and making changes to components. In Table \ref{Table2}, ``Partial Freeze (PF)" indicates freezing some layers (input embedding layer and feature extraction layer) of the TSPN. This is to analyze the impact of free adjustment of other layers on performance while retaining some feature extraction capabilities. In order to analyze the impact of temperature parameter ``$T$=1.0" and scheduled sampling rate ``$\epsilon$=0.5" on the RFRLF method, we also set $T$ in \{0.5,2.0\} and $\epsilon$ in \{0.3,0.7\} , where $T$ is the temperature parameter during data collection and $\epsilon$ is the scheduled sampling rate after the TSPN training epoch is greater than 2. These parameters are selected to observe the impact of smaller and larger temperatures and sampling rates on model exploration and training stability.

\begin{table}[htbp]
	\centering
	\caption{Evaluation Results of Different Agents, Including Maximum, Average, and IQM Metrics.}
	\label{Table2}
	\begin{tabular}{p{1.5cm}<{\centering}p{1.2cm}<{\centering}p{1.2cm}<{\centering}p{1.2cm}<{\centering}p{1.2cm}<{\centering}}
		\toprule
		\textbf{Method} & \textbf{Max Value} & \textbf{Mean Value} & \textbf{Original IQM} & \textbf{Normal IQM} \\
		\midrule
		RFRLF & 3.97$\times$10\textsuperscript{5} & 3.29$\times$10\textsuperscript{5} & 3.64$\times$10\textsuperscript{5} & 0.91 \\ [0.2cm]
		T=0.5 & 3.75$\times$10\textsuperscript{5} & 3.12$\times$10\textsuperscript{5} & 3.43$\times$10\textsuperscript{5} & 0.90 \\[0.2cm]
		T=2.0 & 3.89$\times$10\textsuperscript{5} & 3.33$\times$10\textsuperscript{5} & 3.67$\times$10\textsuperscript{5} & 0.93 \\[0.2cm]
		$\epsilon=0.3$ & 3.77$\times$10\textsuperscript{5} & 3.16$\times$10\textsuperscript{5} & 3.48$\times$10\textsuperscript{5} & 0.91 \\[0.2cm]
		$\epsilon=0.7$ & 3.93$\times$10\textsuperscript{5} & 3.26$\times$10\textsuperscript{5} & 3.57$\times$10\textsuperscript{5} & 0.88 \\[0.2cm]
		PF & 3.91$\times$10\textsuperscript{5} & 3.25$\times$10\textsuperscript{5} & 3.57$\times$10\textsuperscript{5} & 0.90 \\
		\bottomrule
	\end{tabular}
\end{table}


Overall, the performance of RFRLF is comparable to the baseline with altered hyperparameters and outperforms the baseline with partial component structure changes in terms of Mean Value and IQM. Specifically, by comparing with the ``PF" baseline, it is found that after freezing part of the TSPN, the network performance does not change much. The frozen input embedding and feature extraction layers preserve the network's feature extraction capabilities, while the unfrozen components remain adaptable during training, collectively ensuring the effectiveness of the method.

For the temperature parameter, the ``$T$=2.0" baseline performs best. This is because when training the TSPN, a higher temperature coefficient enhances exploration, enabling the agent to explore more states, thereby providing the TSPN with richer data. However, as the temperature coefficient increases, instability also increases, which means that the $T$ should not be too high, and a balance needs to be found between exploration and stability. Experimental results show that, ``$T$=1.0" as the default value is a suitable compromise. In addition, by comparing the ``$T$=0.5", ``$T$=2.0", ``$\epsilon$=0.3", and ``$\epsilon$=0.7" baselines, we find that hyperparameters have a certain impact on the performance of RFRLF, but in general, RFRLF is not very sensitive to changes in hyperparameters. This shows that RFRLF can adapt to different hyperparameter settings.

\section{Conclusion}\label{5}
This paper proposes a novel Reward-Free Reinforcement Learning Framework (RFRLF) that optimizes decision-making processes without requiring manually designed reward signals or explicit action information, through a dual-module architecture comprising the Target State Prediction Network (TSPN) and the Reward-Free State-Guided Policy Network (RFSGPN). The TSPN predicts the next state based on the agent's current state and executed action, while the RFSGPN leverages the prediction results and expert demonstration data (without action information) for policy learning. Experimental results demonstrate that RFRLF outperforms multiple reward-based and reward-free reinforcement learning baselines in Carla and Autocar simulation environments, achieving significant advantages in maximum, average, and IQM metrics. Furthermore, in real-world validation on autonomous driving tracks, the framework successfully achieves path-following tasks without manually designed reward signals, demonstrating its effectiveness in practical scenarios.

Although RFRLF demonstrates strong performance in both simulated and real-world environments, its effectiveness in complex scenarios (e.g., extreme lighting conditions, dynamic obstacles) requires further improvement. Future research will explore state prediction enhancement methods based on multi-modal perception and extend the framework to multi-agent policy optimization scenarios, thereby improving performance in complex environments.

\bibliographystyle{IEEEtran}
\bibliography{ref} 

\end{document}